%% file: paper.tex
\documentclass[11pt,a4paper]{article}
\usepackage[hyperref]{acl2020}
\usepackage{times}
\usepackage{latexsym}
\usepackage{subfigure}
\usepackage{graphicx}

\usepackage{csquotes}

\usepackage{multirow}
\usepackage[normalem]{ulem}
\useunder{\uline}{\ul}{}
\usepackage{booktabs}

\usepackage{afterpage}
\usepackage{boldline}
\usepackage{cleveref}
\crefname{section}{§}{§§}
\Crefname{section}{§}{§§}

\usepackage[ruled, noend,vlined,boxed,commentsnumbered]{algorithm2e}

\usepackage{microtype}

\aclfinalcopy % Uncomment this line for the final submission
\newcommand\model{{\sc Rapformer}}

\title{Rapformer: Conditional Rap Lyrics Generation \\ with  Denoising Autoencoders}
\author{Nikola I. Nikolov$^\dagger$, Eric Malmi$^\ddagger$, Curtis G. Northcutt$^\mathsection$, Loreto Parisi$^\diamond$\\
   $^\dagger$Institute of Neuroinformatics,
  University of Zurich and ETH Zurich  \\ %\hspace{0.2cm} 
  $^\ddagger$Google  \hspace{0.2cm} $^\mathsection$MIT  \hspace{0.2cm} $^\diamond$Musixmatch \\
  \texttt{niniko@ini.ethz.ch}\hspace{0.5cm}
  \texttt{emalmi@google.com} 
   \hspace{0.5cm} \\ \texttt{cgn@mit.edu}
   \hspace{0.5cm}
  \texttt{loreto@musixmatch.com}
  } 

\date{}

\begin{document}
\maketitle
\begin{abstract}

The ability to combine symbols to generate language is a defining characteristic of human intelligence, particularly in the context of artistic story-telling through lyrics. We develop a method for synthesizing a rap verse based on the content of any text (e.g., a news article), or  for augmenting pre-existing rap lyrics. Our method, called \model{}, is based on training a Transformer-based denoising autoencoder to reconstruct rap lyrics from content words extracted from the lyrics, trying to preserve the essential meaning, while matching the target style. \model{} features a novel BERT-based paraphrasing scheme for rhyme enhancement which increases the average rhyme density of output lyrics by $10\%$. Experimental results on three diverse input domains show that \model{} is capable of generating technically fluent verses that offer a good trade-off between content preservation and style transfer. Furthermore, a Turing-test-like experiment reveals that \model{} fools human lyrics experts 25\% of the time.\footnote{We created two songs with lyrics generated by \model{}: using the abstract of this paper as input (see the suppl. mat., and \url{https://bit.ly/37ekn6i}), and using blog posts on AI and creativity as input, video available at \url{https://rapformer.page.link/demo}.}

\end{abstract}

\section{Introduction}

\begin{figure}[!ht]
% \hspace*{-0.4cm} 
    \centering
    \includegraphics[width=1.0\linewidth]{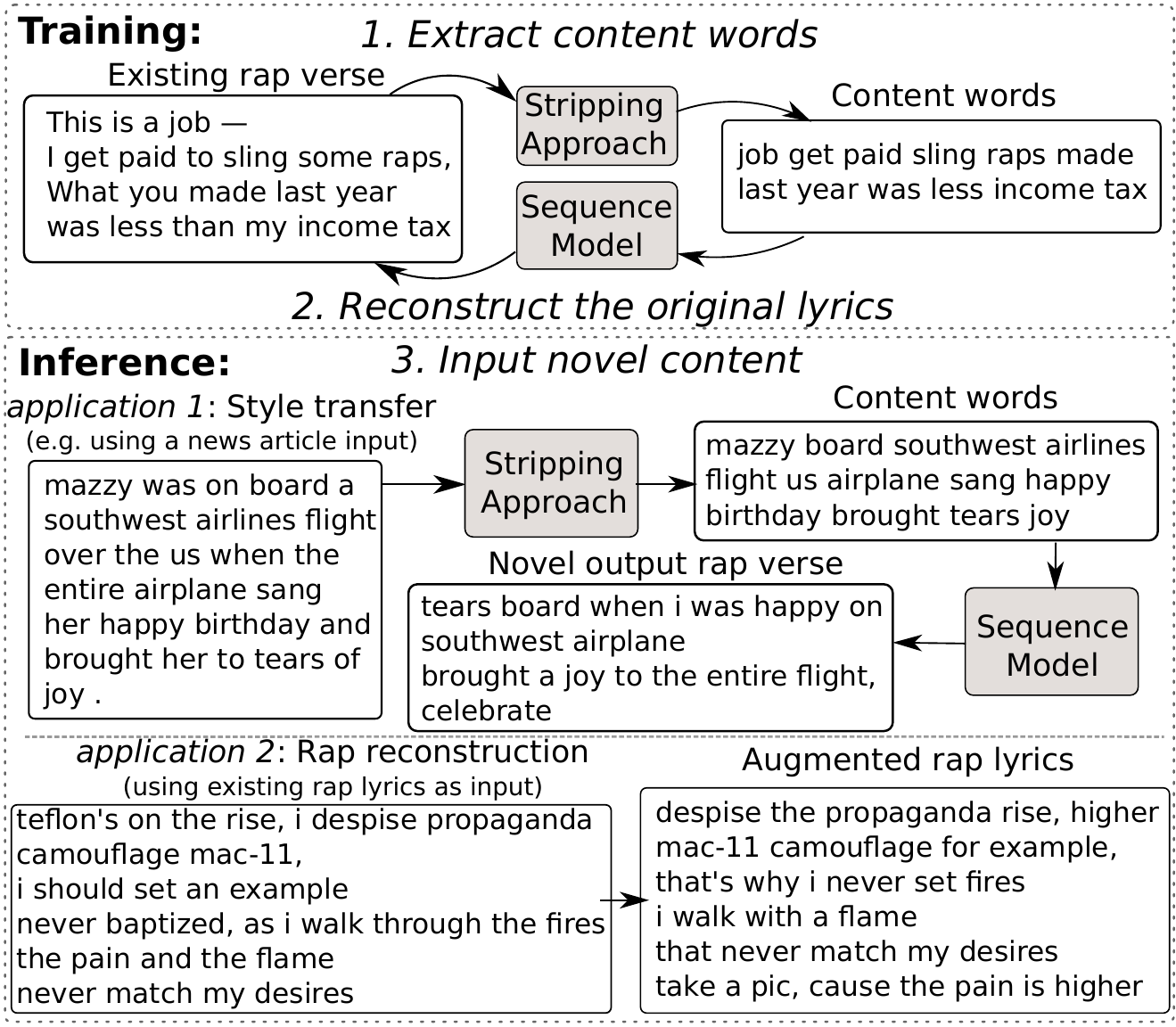}
    \caption{\small{Overview of our approach to \emph{conditional rap lyrics generation}. \textbf{Training}: (1) extract content words from existing rap verses, then (2) train sequence models to guess the original verses conditioned on the content words. \textbf{Inference}: (3) Input content from non-rap texts to produce \emph{content-controlled} rap verses; or input existing rap verses to augment them.}}
    \label{fig:approach}
\end{figure}

Automatic lyrics generation is a challenging language generation task for any musical genre, requiring story development and creativity while adhering to the structural constraints of song lyrics. Here we focus on the generation of \emph{rap lyrics}, which poses three additional challenges specific to the rap genre: ($i$) a verse in rap lyrics often comprises multiple rhyme structures which may change throughout a verse \cite{bradley2017book}, ($ii$) the number of words in a typical rap verse is significantly larger when compared to other music genres \cite{mayer2008rhyme}, requiring modeling of long-term dependencies, and ($iii$) the presence of many slang words.

Prior approaches to rap generation typically use \textit{unconditional} generation \cite{potash2015ghostwriter,malmi2016dopelearning}. That approach synthesizes lyrics without providing any context that could be useful to guide the narrative development into a coherent direction \cite{Dathathri2020Plug}. For example, generating rap lyrics on a specific topic, e.g., "cooking," is not possible with unconditional generation. Motivated by this, in this paper, we propose a novel approach for \textit{conditional} generation of rap verses, where the generator is provided a source text and tasked with transferring the style of the text into rap lyrics. Compared to unconditional generation, this task can support the human creative process more effectively as it allows a human writer to engage with the generator by providing the content of the lyrics while receiving automatic suggestions on how to improve the style of the lyrics to resemble the rap domain.

Our approach to conditional generation is to train sequence-to-sequence models \cite{vaswani2017attention} to reconstruct existing rap verses conditioned on a list of content words extracted from the verses (Figure \ref{fig:approach}). By learning a mapping from content words to complete verses, we implicitly learn the latent structure of rap verses given content, while preserving the target output style of the rap lyrics. Model outputs are enhanced by a post-processing step (Section \ref{sec:rhyme_enhancement}) that substitutes non-rhyming end-of-line words with suitable rhyming alternatives. 

We test our method on three diverse input domains: short summaries of news articles, movie plot summaries, and existing rap lyrics. Automatic and human evaluations (Sections \ref{sec:auto_evaluation} and \ref{sec:human_evaluation}) suggest that our method provides a trade-off between content preservation and style compared to a strong information retrieval baseline. 

\section{Background}

\subsection{Rap Lyrics Generation}

Prior work on rap lyrics generation often focuses on unconditional generation, either using language models \cite{potash2015ghostwriter} or by stitching together lines from existing rap lyrics using information retrieval methods \cite{malmi2016dopelearning}. There are two main drawbacks of unconditional generation of rap lyrics. First, the open-ended nature of the task is too unconstrained for generating lyrics with more specific content: ideally, we may want to have control over at least some aspects of the model during inference, such as the topic of the lyrics, or their sentiment. Second, although frequent rhyming is an essential feature of fluent rap verses \cite{malmi2016dopelearning}, language models have no built-in incentive to learn to consistently generate rhymes at the end of each line, prompting researchers to invent techniques to promote rhyming in their models separately \cite{hopkins2017automatically}.

More recently, \citet{manjavacas2019generation} propose a conditional approach to rap lyrics generation, which extracts high-level features from the lyrics, such as their sentiment, mood, or tense, to provide a template during generation. Although their approach allows for some control during generation, it is limited in terms of generating lyrics with more specific content. The work that is closest to ours is \cite{lee2019neural}
who propose an approach to sentence style transfer based on text denoising, and test their approach on style transfer from pop to rap lyrics. In contrast to these works, we condition the model on longer input text and also introduce a novel method for enhancing the rhymes of our output verses. We also perform extensive automatic and human evaluations on style transfer from diverse input domains to rap lyrics.

\subsection{Text Rewriting and Style Transfer}

Recent work on style transfer of text  \cite{fu2018style,shen2017style,prabhumoye2018style, lample2018multiple, liu2019revision}, focuses on transfer from one text attribute to another, such as gender or political inclination. The main difference between such studies and our work is that our setting is more lenient with respect to meaning preservation: our focus here is on generating creative and fluent verses that match the overall topic of the input and also preserve \textit{some} of the content.
Our conditional lyrics generation based on denoising autoencoders is also related to recent work on self-supervised pre-training objectives for text-to-text generation tasks, which have been beneficial for many NLP tasks, such as automatic text summarization \cite{zhang2019pegasus}, question answering \cite{lewis2019bart,raffel2019exploring}, and data-to-text generation \cite{freitag2018unsupervised}.

\section{Conditional Generation of Lyrics}

Our approach to conditional generation of rap verses consists of three steps (Figure \ref{fig:approach}). 
\begin{enumerate}
\item Given a dataset of rap verses, we apply a stripping approach to extract from each verse a set of \textit{content words} %$\mathbf{c} = \{c_1, ..., c_i\}$ 
that aim to resemble the main content of the original text, omitting any specific stylistic information. 
\item We train a Transformer model \cite{vaswani2017attention} to reconstruct the original rap verses 
conditioned on the content words. 
The model learns to generate the original verse, filling in missing stylistic information.
\item At inference time, we can input content words extracted from a text written in any style, such as a news article, resulting in novel output rhyme verses. After generation, we optionally apply a rhyme enhancement step (Section \ref{sec:rhyme_enhancement}).  
\end{enumerate}

\subsection{Stripping Approach}\label{sec:stripping}

Given a dataset of original rap verses, our base approach to extracting content words involves preprocessing each verse to remove all stop words\footnote{We use the list of English stopwords defined in NLTK.}, numbers, and punctuation. To promote greater novelty\footnote{In early experiments, we tested training models using only this base approach. The models performed very well at reconstructing existing rap lyrics, however when the input was from a different domain, we observed very conservative outputs.} and variability in the outputs produced by our models, we additionally apply one of three noise types to the stripped content words: 

\paragraph{Shuffle.} We shuffle all of the content words on the sentence level (line level for rap verses). This type of noise forces our models to learn to rearrange the location of the input content words when generating the output rap lyric, rather than to merely copy words from the input in an identical order. A similar noising approach has been recently employed by \citet{raffel2019exploring}.

\paragraph{Drop.} We randomly remove 20\% of the input content words for the purpose of promoting generation of novel words, rather than only copying content words from the input. 

\paragraph{Synonym.} We replace 20\% of the content words with synonyms obtained from WordNet \cite{miller1995wordnet}. We pick words randomly and replace them with a random synonym. This type of noise promotes our models to learn to replace content words with synonyms, which might fit better in the context or style of the current output rap verse.

\subsection{Rhyme Enhancement with BERT}\label{sec:rhyme_enhancement}

To improve the rhyming fluency of our models, we implement a post-processing step for \textit{rhyme enhancement (RE)} which modifies a generated verse to introduce additional end-of-line rhymes. Given two lines from a generated verse, such as: 

{\renewenvironment{quote}
  {\small\list{}{\rightmargin=0.5cm \leftmargin=0.5cm}%
   \item\relax}
  {\endlist}
\begin{displayquote}
\small
\vspace{-0.1cm}
\textit{where were \textbf{you}?}\\
\textit{last year i was paid in a drought with no \textbf{beginners}}
\end{displayquote}
\vspace{-0.1cm}
}

RE iterates over each of the lines in the verse, replacing the ending words with a \textit{MASK} token. The verse is then passed through a BERT model\footnote{We finetune a BERT base model on our rap verse dataset for 20 epochs.} \cite{devlin2018bert} which predicts the $K=200$ most likely replacement candidates for \textit{MASK}. For exam ple, the replacement candidates for \textit{you} might be \textit{\{they, we, I, it\}}, and for \textit{beginners} might be \textit{\{food, fruit, you, rules\}}. We pick the candidate that leads to the highest increase in rhyming, determined by the length of the longest overlapping vowels in the two words \cite{malmi2016dopelearning}. In the example above, replacing \textit{beginners} with \textit{food} maximizes the rhyme length, and the example becomes: 

{\renewenvironment{quote}
  {\small\list{}{\rightmargin=0.5cm \leftmargin=0.5cm}%
  \item\relax}
  {\endlist}
\begin{displayquote}
\vspace{-0.1cm}
\small
\textit{where were you?}\\
\textit{last year i was paid in a drought with no \textbf{food}}
\end{displayquote}
\vspace{-0.3cm}
}

\input{algorithms/rhyme_enhancement.tex}

Algorithm \ref{alg:rhyme_enhancement} contains a detailed implementation of our approach. 

\section{Experimental Setup}

\paragraph{Datasets.}

\input{tables/dataset-stat.tex}
 
We conduct experiments using three datasets. As our rap dataset, we use 60k English rap lyrics provided by Musixmatch.\footnote{\url{https://www.musixmatch.com/}}

We split each lyric into verses (in the dataset, each verse is separated by a blank line), remove verses shorter than 4 lines in order to filter for song choruses and intros, and reserve 2k song lyrics for validation and testing. We use two datasets as our out-of-domain inputs: (1) the summaries from the CNN/DailyMail news summarization dataset \cite{hermann2015teaching} and (2) a subset of the CMU movie plot summary corpus \cite{bamman-etal-2013-learning}. Since some of the movie summaries are very long, for this dataset, we filter summaries longer than 140 tokens and shorter than 40 tokens. Table \ref{tab:data-stat} contains detailed statistics of the datasets used for training/validation/testing in our experiments. 

\paragraph{Model details.} As our sequence transducer, we use a 6-layer Transformer encoder-decoder model \cite{vaswani2017attention}. We initially train our models on the source domain (e.g., news articles) for $20$ epochs, after which we finetune them on rap verses for an additional $20$ epochs, using the same stripping approach for both. We train all of our models on the subword level \cite{sennrich-etal-2016-neural}, extracting a common vocabulary of 50k tokens from a joint collection of news summaries and rap lyrics. We use the same vocabulary for both our encoders and decoders and use the Fairseq library.\footnote{\url{https://github.com/pytorch/fairseq}} We train all of our models on a single GTX 1080 Ti card. 

\paragraph{Generation details.}

During inference, we generate outputs using diverse beam search \cite{vijayakumar2016diverse} to promote greater diversity across the hypothesis space. We use a beam with a size of 24 and 6 diverse beam groups. Furthermore, we limit the maximum output sequence length to two times the length of the input content words and penalize repetitions of bigrams in the outputs. 

To select our final output, we additionally implement a simple hypothesis reranking method. For each of the 24 final predictions on the beam, we compute two scores: the rhyme density ($RD$) of the text, following \cite{malmi2016dopelearning}, as well as its repetition score:  
\begin{equation}
\label{eq:repeat}
    rep(\mathbf{s})=\frac{\sum_i \, overlap(\overline{\mathbf s_i}, s_{i})}{|\mathbf{s}|}.
\end{equation}
$rep$ measures the average unigram overlap (see Section 5.1) of each sentence $s_{i}$ in the text $\mathbf{s}$ with all other sentences of the text concatenated into a single string (denoted as $\overline{\mathbf s_i}$). We pick the hypothesis that maximizes: $score(\mathbf{s}) = RD(\mathbf{s}) - rep(\mathbf{s})$. Afterwards, we optionally apply our rhyme enhancement step, to further increase the frequency of rhymes in our outputs.

\paragraph{Bias mitigation} 

Rap lyrics, like other human-produced texts, may contain harmful biases and offensive content which text generation models should not propagate further. Our conditional lyrics generation setup is less susceptible to this issue since the user provides the content, and the generator is supposed to modify only the style of the text. Yet, the model may learn to use inappropriate individual terms that are common in rap lyrics. To alleviate this, we maintain a deny list of words that the model is not able to generate.

\input{tables/automatic-results.tex}

\section{Automatic Evaluation}\label{sec:auto_evaluation}

We conduct an automatic evaluation of \model{}, using the test sets of each of our three datasets. Our focus is on measuring two components that are important for generating fluent conditional rap verses: preserving content from the input text to the output, and maintaining rhyming fluency during generation. 

\subsection{Evaluation Metrics}\label{sec:auto_metrics}

\paragraph{Content preservation.} We test the capacity of our models to preserve content words from the input by computing a unigram overlap score: 
\begin{equation}
\label{eq:overlap}
    overlap(\mathbf{x}, \mathbf{y})
= \frac{|\{\mathbf{y}\} 	\cap \{\mathbf{x}\}|}{|\{\mathbf{y}\}|}
\end{equation}
between unique unigrams from an input text $\mathbf{x}$ and the generated output rap verse $\mathbf{y}$. We also report the BLEU score \cite{papineni-etal-2002-bleu} when training a model to reconstruct original lyrics. 

\paragraph{Rhyming fluency.} We measure the technical quality of our rap verses using the rhyme density (RD) metric \cite{malmi2016dopelearning}.\footnote{\url{https://github.com/ekQ/raplysaattori}} The metric is based on computing a phonetic transcription of the lyrics and finding the average length of matching vowel sound sequences which resemble multisyllabic assonance rhymes. As a reference, RD values above $1$ can be considered high, with some rap artists reaching up to 1.2. 

\subsection{Baselines}

For reference, we report the result of an information retrieval baseline, which retrieves the closest text from our training dataset given input from the news or movies test sets, using sentence embedding similarity.\footnote{We use a 600-dimensional Sent2Vec model \cite{pgj2017unsup}, which is pretrained on Wikipedia.} We report two variants of the IR baseline. First, we retrieve the closest summary from the CNN/DailyMail news training set ({\sc IR News}), which resembles a lower bound for our target task of style transfer from news to rap lyrics. Second, we retrieve the closest verse from our rap training set ({\sc IR Rap}). The outputs of the strong IR Rap baseline perfectly match the style of original rap verses, giving us an upper bound for rap style, while maintaining some degree of lexical and semantic overlap with the input texts. 

\subsection{Results}

Our results are shown in Table \ref{tab:automatic-results-full}, where we include all of our stripping approaches (Shuffle, Drop, Replace). We report the results of applying the additional rhyme enhancement step separately (model names ending with {"+ RE"}). 

\paragraph{Rap reconstruction.}

In the left part of Table \ref{tab:automatic-results-full}, we evaluate our model's capacity to reliably regenerate original rap lyrics given extracted content words from them. \model{} performed well on this task, generating fluent lyrics that incorporate a large part of the input content words and surpassing the average rhyme density observed in the training dataset ({\sc Inputs}). When using our rhyme enhancement step, we observe a slight decrease in overlap due to the potential replacement of content words. However, RD increases by 10\% on average. 

\paragraph{Style transfer.} 

In the right part of Table \ref{tab:automatic-results-full}, we evaluate the capacity of our model to generate rap lyrics using content words extracted from movie plot summaries or news article summaries. For these inputs, our model generated outputs with lower overlap on average than for rap reconstruction, with movies retaining slightly more content than news. This gap is potentially due to the large differences in style, vocabulary, and topic of the inputs, prompting our models to ignore some of the content words to better match the target rap style. Still, our generation methods manage to achieve similar RD scores while considerably outperforming the strong IR baseline in terms of overlap. 

\section{Human Evaluation}\label{sec:human_evaluation}

Due to the limitations of automatic metrics for text generation,
we also perform four human evaluation experiments using three raters, who are trained to translate lyrics. Due to limited resources, we evaluate only the \model{} variant with the {\sc Shuffle} stripping approach and rhyme enhancement, which achieved the highest content overlap in our automatic evaluation. 

\begin{table}[ht]
	\centering
	\small
	\setlength\tabcolsep{3pt}
    \begin{tabular}{c|ccc}
    \toprule
     \textbf{Method}  & \textbf{Style} & \textbf{Meaning} & \textbf{Familiarity} \\
     \midrule
    {\sc IR News} & 1.18 & 2.01 & 1\% \\ 
     {\sc IR Rap} & 4.27 & 1.33 & 31\% \\  
     \model{} & 2.03 & 2.55 & 8\% \\
     \bottomrule
    \end{tabular}
    \caption{Human evaluation results of \model{} (using the {\sc Shuffle} stripping approach, and news articles as input). The average inter-rater agreement for \textbf{Style} is $0.3$, and for \textbf{Meaning} is $-0.1$, measured using Cohen's Kappa \cite{cohen1960coefficient}. }
    \label{tab:human_eval_style}
    
\end{table}

The first two human experiments (in Table \ref{tab:human_eval_style}) focus on style transfer using news articles as input. Each rater inspected 100 verses produced by either the \model{}, or the two IR baselines, answering the following three questions:

\begin{enumerate}
    \item \textit{How much do the lyrics presented resemble rap lyrics? On a scale from 1 (not at all), to 5 (this could be from existing rap lyrics)}, which measures the capacity of our models to preserve the \textbf{Style}.
    \item \textit{How well do the lyrics preserve the content of the original news article on a scale from 1 (not at all) to 5 (very well)?} This question measures the meaning preservation of our models (\textbf{Meaning}).
    \item \textit{Do these lyrics look like a song you know (yes or no)?} For {\sc IR Rap}, this question measures the \textbf{Familiarity} of the raters with the lyrics; for the other two methods, it measures the capacity to fool the raters. 
\end{enumerate}

\begin{table}[ht]
	\centering
	\small
	\setlength\tabcolsep{3pt}
    \begin{tabular}{c|cc}
    \toprule
     \textbf{Method}  & \textbf{Side-by-Side} & \textbf{Random} \\
     \midrule
     \model{} & 7\% & 25\% \\
     \bottomrule
    \end{tabular}
    \caption{Turing-like evaluation, reporting the percentage of lyrics generated by \model{} (using the {\sc Shuffle} stripping approach, and rap lyrics as input) that human experts incorrectly label as existing rap lyrics. The average inter-rater agreement for \textbf{Side-by-Side} is $0.8$, and for \textbf{Random} is $0.4$, measured using Cohen's Kappa \cite{cohen1960coefficient}. }
    \label{tab:human_eval_turing}
    
\end{table}

The other two human experiments (in Table \ref{tab:human_eval_turing}) focus on our rap reconstruction task, performing two Turing-test-like comparisons between 100 real and synthetic verses: 

\begin{enumerate}
    \item \textbf{Side-by-Side}: the original rap lyrics and \model{} lyrics are presented side-by-side, in a random order, and a rater is asked, \emph{Which of these lyrics was written by a human?} (see the Appendix for examples). 
    \item \textbf{Random}: a verse is shown and the rater is asked, \emph{"Do you think these rap lyrics are: (a) AI-generated or (b) human-created?"}. 
\end{enumerate}

In terms of style (Table \ref{tab:human_eval_style}), we outperform {\sc IR News}, demonstrating that there is a change in style towards rap verses. There is still a large gap to reach the fluency of original rap verses retrieved by {\sc IR Rap}. However, it is worth noting that the content preservation of {\sc IR Rap} is considerably lower, as shown in Tables~\ref{tab:automatic-results-full} and \ref{tab:human_eval_style}, and simply the fact that the content of the generated lyrics is closer to the news domain might encourage the raters to rate the generated lyrics as having a lower rap resemblance score. In other words, the style score of {\sc IR Rap} might be unrealistic to attain even with a perfect conditional generator. 

\input{tables/examples_rap}
\input{tables/examples_movies}
\input{tables/examples_news}
\input{tables/turing-examples-wide}

Overall, the results indicate that our method provides a trade-off between the two baselines in terms of style while outperforming them in terms of content overlap. Furthermore, 8\% of the time, our conditional generation model fooled experienced raters to think that our synthetic rap lyrics generated from news articles originate from real rap songs. Our rap lyrics augmentation approach also proved to be robust in the Turing-style evaluation of rap reconstruction (Table \ref{tab:human_eval_turing}), where \model{} fooled the raters 25\% of the time when lyrics from a random source are presented one-by-one, and 7\% of the time when lyrics are presented side-by-side. 

\section{Example Model Outputs}\label{sec:ex_ouptuts}

In Tables \ref{tab:examples_rap}, \ref{tab:examples_movies} and \ref{tab:examples_news}, we also display a few manually selected example model outputs (additional examples are available in the Appendix) produced after inputting content words extracted from each of our input text styles (existing rap lyrics, movie plot summaries and news article summaries). When using existing rap lyrics as input, many outputs seem coherent and of higher quality in comparison to outputs produced using news/movie inputs. For news/movie inputs, the models are still capable of integrating the input content words into a rhyming verse that preserves some of the overall meaning of the original text (e.g., "the film also follows the adventures of lucius the slave escaping via the underground railroad to freedom" $\rightarrow$ "slave, run from lucius slavery; battle of freedom and liberty"). 

Furthermore, in Table \ref{tab:examples_turing} we present examples from our side-by-side Turing test, where we asked raters to choose which of two lyrics was generated (augmented) by \model{}, and which was written by a human. For the selected outputs, two of the three raters incorrectly guessed that the lyrics generated by \model{} were actually human-created. 

\section{Conclusion}

We have proposed a novel approach to generation of rap verses conditioned on a list of content words. We showed that our method is capable of generating coherent and technically fluent synthetic verses using diverse text types as input, including news articles, movie plot summaries, or original rap verses. The fluency of our outputs is further improved through a novel rhyme enhancement step. Our approach is particularly effective when rephrasing the content of existing rap lyrics in novel ways, making it a potentially useful tool for creative writers wishing to explore alternative expressions of their ideas. 

The generality of our approach to conditional text generation makes it applicable to generation of creative texts in other domains, such as poetry or short stories. Future work could explore other approaches to extracting content words, including combining several stripping approaches, and could explore the utility of large-scale pretrained models (e.g., \citep{raffel2019exploring, lewis2019bart}) for this task. Another direction is to extend our work to end-to-end generation with an integrated rhyming loss function, which could potentially be tackled using reinforcement learning \cite{Luo19DualRL}. Moreover, the task of generating coherent lyrics from a set of content words could be naturally modeled as a text-editing task \cite{dong2019editnts,mallinson2020felix,malmi2019encode} instead of a sequence-to-sequence task. 

\section*{Acknowledgements}

We are thankful to Alessandro Calmanovici, Scott Roy, Aliaksei Severyn, and Ada Wan for useful discussions. We also thank Simone Francia and Maria Stella Tavella from Musixmatch, for technical help, and the three raters, for participating in the human evaluation.

\bibliography{paper}
\bibliographystyle{acl_natbib}

\appendix

\newpage 

\section{Additional Model Outputs}\label{sec:examples_appendix}

In Tables \ref{tab:extra_examples_rap}, \ref{tab:extra_examples_style_transfer_movies} and \ref{tab:extra_examples_style_transfer_news} we display a few additional manually selected model outputs for each of our input domains (rap lyrics, movie summaries and news article summaries) and each of our stripping approaches ({\sc Shuffle} (\model{}), {\sc Drop}, and {\sc Synonym}).

\section{Demo Song}\label{sec:demo_song}

We generated lyrics for a demo song by using the abstract of this paper as the input to \model{}. We generated multiple samples, by reshuffling the content words of the abstract multiple times. We sent all sample lyrics to a rap artist, and asked them to record a song using a subset of those lyrics. We allowed for re-arranging and deletion, but no addition of human-created lyrics. The resulting audio file is included in the supplementary material \footnote{Also available at \url{https://bit.ly/37ekn6i}.}, while the final lyrics of the song are in Table \ref{tab:abstract_lyrics}. 

We also tested the recently released Jukebox algorithm \cite{dhariwal2020jukebox} for end-to-end synthesis of a rap song conditioned on the abstract lyrics. However, our preliminary results were unsatisfactory since it was impossible to tell individual words apart from the generated audio.

\input{tables/appendix_examples}

\input{tables/abstract-lyrics}

\end{document}

%% file: algorithms/rhyme_enhancement.tex
\begin{algorithm}[t]
 \caption{Bert Rhyme Enhancement}  \label{alg:rhyme_enhancement}

\small 
\DontPrintSemicolon
\SetKwData{Verse}{$\mathbf{V}$}
\SetKwData{K}{$K$}
\SetKwData{Index}{$i$}
\SetKwData{FirstLine}{$source\_line$}
\SetKwData{SecondLine}{$target\_line$}
\SetKwData{FirstWord}{$src$}
\SetKwData{SecondWord}{$tgt$}
\SetKwData{RLOriginal}{$rl\_orig$}
\SetKwData{RLNew}{$rl\_new$}
\SetKwData{RLFirst}{$rl\_1$}
\SetKwData{RLSecond}{$rl\_2$}
\SetKwData{MaskedFirst}{$mask\_1$}
\SetKwData{MaskedSecond}{$mask\_2$}
\SetKwData{FirstCandidate}{$cand\_1$}
\SetKwData{SecondCandidate}{$cand\_2$}

\SetKwData{Predictions}{$preds$}
\SetKwData{Up}{up}
\SetKwFunction{BERT}{bert\_predictions}
\SetKwFunction{RL}{rhyme\_length}
\SetKwFunction{MaskText}{mask\_text}
\SetKwInOut{Input}{input}
\SetKwInOut{Output}{output}
 \Input{lyrics verse \Verse $= \{l_0,..., l_N\}$ consisting of $N$ tokenized lines; number of BERT predictions \K to consider.}
 \Output{modified \Verse with enhanced rhyming.}
 \BlankLine
\SetKwFunction{FMain}{get\_rhyming\_replacement}
\SetKwProg{Fn}{Function}{:}{}
\Fn{\FMain{\Verse, src\_idx, tgt\_idx, mask}}{
    \FirstWord $\leftarrow$ \Verse[src\_idx][-1] \tcp{get last word}
    \SecondWord $\leftarrow$ \Verse[tgt\_idx][-1]\;
    \BlankLine

    \tcp{Predict most likely words.}
    \Predictions $\leftarrow$ \BERT($mask$, $K$)\;
    \tcp{Compute original rhyme length.}
    \RLOriginal $\leftarrow$ \RL(\FirstWord, \SecondWord)\;
    \For{$pred$ $\in$ $\Predictions$}{  
        \RLNew $\leftarrow$ \RL($pred$, \SecondWord)\;
        \uIf{\RLNew $>$  \RLOriginal 
        }{
            \tcp{return replacement}
            \KwRet $pred$, \RLNew\; 
        } 
    }
    \KwRet $target$, \RLOriginal \tcp{return original} 
}
\;
 \BlankLine
\For{\Index$\leftarrow 1, 3,...,N$ \tcp{for each odd line}}{
    \tcp{Create two masks for the two consecutive lines.}
    \MaskedFirst $\leftarrow$ \MaskText(\Verse, i)\;
    \MaskedSecond $\leftarrow$ \MaskText(\Verse, i + 1)\;
    \tcp{Generate replacement candidates.}
    \FirstCandidate, \RLFirst $\leftarrow$ \FMain(\Verse, \Index + 1, \Index, \MaskedFirst) \tcp{replace last word at \Index}
    \SecondCandidate, \RLSecond $\leftarrow$ \FMain(\Verse, \Index, \Index + 1, \MaskedSecond) \tcp{replace last word at \Index + 1}
    \uIf{\RLSecond $\geq$  \RLFirst \tcp{update lines in \Verse}
    }{
        \Verse[\Index + 1][-1] $\leftarrow$ \SecondCandidate 
    }\Else{
        \Verse[\Index][-1] $\leftarrow$ \FirstCandidate\;
    }
    
}
\BlankLine
\KwRet \Verse
 
\end{algorithm}

%% file: tables/dataset-stat.tex
\begin{table}[ht]
\centering
\small
\begin{tabular}{c|c|c|c}
\hlineB{2}
 & \textbf{News} & \textbf{Movies} & \textbf{Rap} \\ \hlineB{2}
 \textit{\# Pairs} & 287k/11k/11k & - / - /12k & 165k/1k/1k \\ \hline
\textit{Sent. p.d.} & 3.7 $\pm$ 1.2 & 3.9 $\pm$ 1.6 & 10.5 $\pm$ 4.5 \\ \hline
\textit{Tok. p.d.} & 57.9 $\pm$ 24.3 & 90 $\pm$ 27.6 & 91.8 $\pm$ 49.1 \\ \hline
\textit{Tok. p.s.} & 15.1 $\pm$ 4.7 & 22.4 $\pm$  11 & 9.5 $\pm$ 4.25 \\ \hlineB{2}
\end{tabular}
\caption{Statistics of our datasets. \textit{\# Pairs} denotes the number of pairs used for training/validation/testing; \textit{p.d.} is per document; \textit{p.s.} is per sentence.}
\label{tab:data-stat}
\end{table}

%% file: tables/automatic-results.tex
\begin{table*}
\centering
\small 
\vspace{-0.1cm}
\begin{tabular}{rl|ccc|cc|cc}
\hlineB{2}
&  & \multicolumn{3}{c|}{\textbf{Rap reconstruction}} & \multicolumn{2}{c|}{\textbf{Style transfer from movies}} & \multicolumn{2}{c}{\textbf{Style transfer from news}} \\ \hline
& \multicolumn{1}{c|}{\textbf{Model}} & \textbf{BLEU} & \textbf{Overlap} & \textbf{RD} & \textbf{Overlap} & \textbf{RD} & \textbf{Overlap} & \textbf{RD} \\ \hlineB{2}
& \multicolumn{1}{c|}{{\sc Inputs}} & - & - & 0.84 $\pm$ 0.38 & - & 0.73 $\pm$ 0.2 & - & 0.72 $\pm$ 0.21 \\ \hlineB{2}
& \multicolumn{1}{c|}{{\sc IR News}} & - & - & - & - & - & 0.29 $\pm$ 0.09 & 0.74 $\pm$ 0.19 \\ %\hline
& \multicolumn{1}{c|}{{\sc IR Rap}} & - & - & - & 0.19 $\pm$ 0.06 & \textbf{1.02 $\pm$ 0.23} & 0.17 $\pm$ 0.06 & 1.01 $\pm$ 0.24 \\ \hlineB{2}
\parbox[t]{0.5mm}{\multirow{6}{*}{\rotatebox[origin=c]{90}{\model}}} & \multicolumn{1}{c|}{{\sc Shuffle}} & 10.27 & \textbf{0.63 $\pm$ 0.13} & 1.01 $\pm$ 0.31 & \textbf{0.51 $\pm$ 0.11} & 0.90 $\pm$ 0.23 & \textbf{0.45 $\pm$ 0.12} & 0.89 $\pm$ 0.26 \\ %\hline
& \multicolumn{1}{c|}{{\sc Shuffle + RE}} & 12.72 & 0.60 $\pm$ 0.12 & 1.10 $\pm$ 0.32 & 0.49 $\pm$ 0.10 & 0.96 $\pm$ 0.27 & 0.43 $\pm$ 0.11 & 0.98 $\pm$ 0.27 \\ %\hline
& \multicolumn{1}{c|}{{\sc Drop}} & 11.06 & 0.52 $\pm$ 0.11 & 1.03 $\pm$ 0.32 & 0.43 $\pm$ 0.10 & 0.90 $\pm$ 0.24 & 0.38 $\pm$ 0.10 & 0.93 $\pm$ 0.25 \\ %\hline
& \multicolumn{1}{c|}{{\sc Drop + RE}} & 09.81 & 0.50 $\pm$ 0.11 & \textbf{1.13 $\pm$ 0.33} & 0.40 $\pm$ 0.09 & 0.99 $\pm$ 0.27 & 0.36 $\pm$ 0.10 & 1.03 $\pm$ 0.26 \\ %\hline
& \multicolumn{1}{c|}{{\sc Replace}} & \textbf{14.30} & 0.57 $\pm$ 0.15 & 1.00 $\pm$ 0.30 & 0.43 $\pm$ 0.14 & 0.86 $\pm$ 0.28 & 0.34 $\pm$ 0.13 & 0.95 $\pm$ 0.27 \\ %\hline
& \multicolumn{1}{c|}{{\sc Replace + RE}} & 12.72 & 0.54 $\pm$ 0.15 & 1.10 $\pm$ 0.31 & 0.40 
$\pm$ 0.13 & 0.98 $\pm$ 0.24 & 0.31 $\pm$ 0.12 & \textbf{1.05 $\pm$ 0.28} \\ \hlineB{2}
\end{tabular}
\caption{Automatic metric results of \model{}, using three alternative stripping approaches: {\sc Shuffle, Drop} and {\sc Replace}. Model names ending in {\sc * + RE} denote use of the additional rhyme enhancement step (see Section \ref{sec:rhyme_enhancement}). {\sc Input} measures the result of the original input texts, for each of the three inputs (rap/movies/news). \textbf{Overlap} is the content preservation score, \textbf{RD} is the rhyme density metric. The highest results for each column are in bold. }
\label{tab:automatic-results-full}
\end{table*}

%% file: tables/examples_rap.tex
\begin{table}
    \centering
		\small 
		\begin{tabular}{p{7.3cm}} %{7.5cm}
    \hlineB{2} 
    {\sc Input (rap): } before the dough came , my whole aim , was blow like propane \newline
control the whole domain , and then show no shame \newline
make rappers go ? and they so lame , playing with no game \newline
put em on the lil plane til they can 't claim no fame \newline
i got , the range , better , stay in the slow lane \newline
i make the flow change from hurricanes to a slow rain \newline
your thoughts are so plane , i rearrange your whole frame \newline
until my whole name grow out your brain like rogaine \newline
...
    \\ \hline
   {\sc Shuffle: } aim dough like propane came with a whole blow \newline shame how you control the whole show \newline lame rappers playin' the game, make your domain go \newline can't claim em til you put a lil fame on a plane \newline slow down, stay in your lane, got a better range \newline make it rain, change your flow, slow down \newline rearrange your whole frame, plane thoughts \newline grow like rogaine, put my whole name on your brain \newline ...  \\ \hline
   {\sc Shuffle + RE: } aim dough like propane came with a whole blow \newline shame how you control the whole show \newline lame rappers playin' the game, make your domain go \newline can't claim em til you put a lil fame on a \textbf{show} \newline slow down, stay in your lane, got a better range \newline make it rain, change your flow, slow \textbf{rain} \newline rearrange your whole frame, plane \textbf{change} \newline grow like rogaine, put my whole name on your brain \newline ...
   \\ \hlineB{2}
    \end{tabular}
	\caption{Example model output for rap reconstruction. Words replaced by our rhyme enhancement step are in bold. The input lyrics are from the song \textit{How I Get Down} by Rakim.}
	\label{tab:examples_rap}
\end{table}

%% file: tables/examples_movies.tex
\begin{table}
    \centering
		\small 
		\begin{tabular}{p{7.5cm}} %{7.5cm}
    \hlineB{2} 
   {\sc Input (movies): } the film follows the lives of several west point cadet classmates who find themselves on opposite sides of the war . the film also follows the adventures of lucius the slave escaping via the underground railroad to freedom with the film cutting between the first battle of bull run and the birth of a lucius ' child born in slavery . \\ \hline
   {\sc Shuffle: } this is the opposite of war follows lives on both sides \newline several point film from the west to the wrong \newline find a child born escaping via film \newline film the underground cutting off the film of all the complications \newline slave, run from lucius slavery \newline battle of freedom and birth \newline also the first bull follows luc-up! \\ \hline
   {\sc Shuffle + RE: } this is the opposite of war follows lives on both sides \newline several point film from the west to the \textbf{light} \newline find a child born escaping via \textbf{immigration} \newline film the underground cutting off the film of all the complications \newline slave, run from lucius slavery \newline battle of freedom and \textbf{liberty} \newline also the first bull follows luc-up! \\ \hlineB{2}
    \end{tabular}
	\caption{Example model outputs for style transfer from movie plot summaries. Words replaced by our rhyme enhancement step are in bold.}
	\label{tab:examples_movies}
\end{table}

%% file: tables/examples_news.tex
\begin{table}
    \centering
		\small 
		\begin{tabular}{p{7.5cm}} 
    \hlineB{2} 
   {\sc Input (news): } temperatures dipped into the mid-30s during 4 days man lay in woods of philadelphia park . mom told police son was with her in maryland , but he was found friday with blanket , bible . victim being treated for malnutrition , dehydration ; mother faces host of charges after extradition . \\ \hline
   {\sc Shuffle: } man i was dipped up in a lay up with some of them from an old \newline mid-30s days in the park \newline mom told me to be in michigan woods \newline police blanket friday \newline i found my son a bible \newline he was a host for the charges \newline my mother treated him as an age \newline a victim of faces \\ \hline
   {\sc Shuffle + RE: } man i was dipped up in a lay up with some of them from an old \newline mid-30s days in the \textbf{home} \newline mom told me to be in michigan anyway \newline police blanket \textbf{friday} \newline i found my son a bible \newline he was a host for the \textbf{trial} \newline my mother treated him as an \textbf{alien} \newline a victim of faces \\ \hlineB{2}
    \end{tabular}
	\caption{Example model outputs for style transfer from news articles. Words replaced by our rhyme enhancement step are in bold.}
	\label{tab:examples_news}
\end{table}

%% file: tables/turing-examples-wide.tex
\begin{table*}[htb]
    \centering
    \resizebox{\textwidth}{!}{%
		\small 
		\begin{tabular}{p{7.5cm} p{7.5cm}} %{7.5cm}
    \toprule
   \textit{Question 45 of 100}\\
{\sc LYRICS (a)} & {\sc LYRICS (b)} \\
waka waka: & i say na correct eye i take waka this waka \\
they say na blind eye, take it far & but after i've got you i blind pata pata\\
i've got it on my own, my own & oche du no dum no oda du num doka\\
oche num, oda du, doka dum so & anybody try you i go shoot the murderfker\\
if anybody ever try go shoot the almighty & ever blazing you amazing\\
blazing so amazing & \\[0.5em]
\textit{Which of these lyrics was written by a human?} & \textit{Correct answer:} {\sc (b)}\\ \midrule

\textit{Question 72 of 100} & \\
{\sc LYRICS (a)} & {\sc LYRICS (b)} \\
vegas on the third floor, like lamar with the cardio & out in vegas like lamar, third floor tropicana\\
fascinated by the cars smokin' dope in the casino & fascinated with the cars, smokin' dope in the phantom\\
despise the propaganda rise, higher & teflon's on the rise, i despise propaganda\\
mac-11 camouflage for example, that's why i never set fires & camouflage mac-11, i should set an example\\
i walk with a flame that never match my desires & never baptized, as i walk through the fires\\
take a pic, cause the pain is higher & the pain and the flame never match my desires\\
i'm rich as a coupe, light it up with kelly & crucified cause i'm rich, in the coupe, take a pic\\
phone sucker, my friend, it's a blessing & on the phone at the light, kelly rowland's a friend\\
benz, plaques, wall, and g6's & catfish in the benz, manti teo's a sucker\\
- 'em all, hustler say the victim & plaques on the wall, hustler so i can say "- 'em"\\
ciroc and bel air - & bel air for the -, ciroc in the pool\\
april -'s -, her name so & my - is a -, her name is april's a fool\\[0.5em]
\textit{Which of these lyrics was written by a human?} & \textit{Correct answer:} {\sc (b)}\\ \midrule

\textit{Question 74 of 100} & \\ 
{\sc LYRICS (a)} & {\sc LYRICS (b)} \\
she cut the call when she was on ma phone & i picked up the phone and cut the line and call\\
when you picked up the line & i asked what's up girl, why you got so long\\
you got so mad and asked me who's the girl & i'm sleeping behind you\\
i'm sleeping with behind & baby, i guess i try to say the truth\\
baby, i had no words to say & but... it's time to lie...\\
so i guess i will try & \\
not to lie... it's the time... & \\[0.5em]
\textit{Which of these lyrics was written by a human?} & \textit{Correct answer:} {\sc (a)}\\ \bottomrule
    \end{tabular}
    }
	\caption{Examples of lyrics generated by \model{} that fooled the majority (at least two out of three) human raters in a side-by-side comparison with human created lyrics. Inappropriate words are replaced by a single dash.}
	\label{tab:examples_turing}
\end{table*}

%% file: tables/appendix_examples.tex
\begin{table*}
    \centering
		\small 
		\begin{tabular}{p{16cm}} %{7.5cm}
    \hlineB{2} 
    {\sc Input: } control the whole domain , and then show no shame \newline
make rappers go ? and they so lame , playing with no game \newline
put em on the lil plane til they can 't claim no fame \newline
i got , the range , better , stay in the slow lane \newline
i make the flow change from hurricanes to a slow rain \newline
your thoughts are so plane , i rearrange your whole frame \newline
until my whole name grow out your brain like rogaine \newline
...
    \\ \hline
    {\sc Drop + RE: } \newline like propane in control, my whole objective show \newline no shame, i can make it playin' the game \newline put em in a lil' game, til they don't claim fame range \newline you better stay in your lane, and make the flow change \newline i be on a plane to rearrange your own frame \newline your whole name grow in my brain, rogaine letter 
  \\ \hline 
  {\sc Synonym + RE: } control the whole domain just to show it no shame \newline make these rappers go lame, they playin' the game \newline put em on a lil' plane til they can't title the fame \newline if you got a range, you better stay in a slow lane \newline name run change hurricane slow rain \newline thoughts on the ground and rearrange your whole frame \newline whole name grow brain like a rogaine cane \\ \hline
  {\sc Shuffle + RE (\model{}): } aim dough like propane came with a whole blow \newline shame how you control the whole show \newline lame rappers playin' the game, make your domain go \newline can't claim em til you put a lil fame on a show \newline slow down, stay in your lane, got a better range \newline make it rain, change your flow, slow rain \newline rearrange your whole frame, plane change \newline grow like rogaine, put my whole name on your brain \newline ... \\ \hlineB{2}
    {\sc Input: } now comes our turn \newline we step in, you burn \newline you got something to learn \newline and bring a breath of fresh air \newline this is how we go \newline you check this mode \newline ride da best you know \newline and still want more
    \\ \hline
    {\sc Drop + RE: } when it comes to turn to step \newline i got something to bring in my breath \newline go check my mode \newline ride with da so \newline you know i still want to be your own 
  \\ \hline 
  {\sc Synonym + RE: } when it comes to turn my step to burn \newline i got something to learn \newline bring your breath to new light \newline go check the way i ride \newline da best i know \newline you still want so \\ \hline
  {\sc Shuffle + RE (\model{}): } when it comes to my turn \newline step up and learn something to burn \newline i got fresh air on my breath \newline go check da best mode, bring it yeah \newline you know i still want a ride \\ \hlineB{2}
    \end{tabular}
	\caption{Additional model outputs for rap reconstruction.}
	\label{tab:extra_examples_rap}
\end{table*}

\begin{table*}
    \centering
		\small 
		\begin{tabular}{p{15cm}} %{7.5cm}
    \hlineB{2} 
   {\sc Input: } hoping to improve his financial lot , petty thief hawk chovinski hires a dancing instructor to teach him how to bear himself like a gentleman . his lessons completed , hawk then poses as a european nobleman , intending to trap a wealthy wife . yolande cowles sees through hawk ' s pose but falls in love with him anyway . \\ \hline
   {\sc Drop + RE: } i improve a grizzly lot of petty thief times \newline dancing in the middle of the night \newline i am the man who can teach you how to bear it \newline like a gentleman with diamonds \newline i'm a superheroic, i can be your wife \newline yolande cowles tonight \newline falls in love anyway \\ \hline
   {\sc Synonym + RE: } hoping that you can improve \newline a financial lot of petty use \newline mortarboard chovinski engage \newline dancing with the snake \newline teach her how to settle \newline like a gentleman \newline lessons are shackled by a bullet \newline sit in european \newline imagine \newline in the trap with a wealthy wife \newline yolande hood sees the sky \newline when the pose falls in line \newline anyway, no, not me \\ \hline
   {\sc Shuffle + RE (\model{}): } you teach me petty dancing like bear thief \newline chovinski, intersect, be \newline a lot of financial gentleman hoping he can improve somebody \newline wife, nobleman, the trap is so polished \newline wealthy hawk lessons, european hawk lessons \newline yolande cowles anyway, sees him pose when he says \newline hawk love!  \\ \hlineB{2}
   {\sc Input: } the film follows the lives of several west point cadet classmates who find themselves on opposite sides of the war . the film also follows the adventures of lucius the slave escaping via the underground railroad to freedom with the film cutting between the first battle of bull run and the birth of a lucius ' child born in slavery . \\ \hline
   {\sc Drop + RE: } film of the west point where they can find the opposite sides of ours \newline film also and they will be a slave escaping me from the underground, \newline and we will not be the same if we are not the maker \newline this is a film cutting first bull from birth to child's slaver. \\ \hline
   {\sc Synonym + RE: } film to succeed our lives in several zones \newline our head is the most likely to find our own \newline we are not the same as the other side of ever \newline film also follows adventure \newline the lucius slave, the escaping via underground \newline motorical, freedom, film out \newline first battle bull, then feed him birth \newline golden child, born in order \\ \hline
   {\sc Shuffle + RE (\model{}): } this is the opposite of war follows lives on both sides \newline several point film from the west to the light \newline find a child born escaping via immigration \newline film the underground cutting off the film of all the complications \newline slave, run from lucius slavery \newline battle of freedom and liberty \newline also the first bull follows luc-up!  \\ \hlineB{2}
       \end{tabular}
	\caption{Additional model outputs for style transfer from movie plot summaries to rap lyrics. }
	\label{tab:extra_examples_style_transfer_movies}
\end{table*}

\begin{table*}
    \centering
		\small 
		\begin{tabular}{p{15cm}} %{7.5cm}
    \hlineB{2} 
    {\sc Input (news): } temperatures dipped into the mid-30s during 4 days man lay in woods of philadelphia park . mom told police son was with her in maryland , but he was found friday with blanket , bible . victim being treated for malnutrition , dehydration ; mother faces host of charges after extradition . \\ \hline
   {\sc Drop + RE: } i've been dipped for days, lay in woods \newline in the park with the crook \newline my son from pittsburgh found friday blanket, bible victim \newline i was born to be a victim of my reality \newline with no faces \newline host charges, i had to do it everything \\ \hline
   {\sc Synonym + RE: } dipped in mid-a.t. days, man i dwell in ways \newline UNK 3: gs, park mom told the police son that he was from illegal \newline found him on friday, he had a blanket with a bible \newline a victim of how he treated him as a  \\ \hline
   {\sc Shuffle + RE (\model{}): } man i was dipped up in a lay up with some of them from an old \newline mid-30s days in the home \newline mom told me to be in michigan anyway \newline police blanket friday \newline i found my son a bible \newline he was a host for the trial \newline my mother treated him as an alien \newline a victim of faces \\ \hlineB{2}
    {\sc Input (news): } jordan spieth won the 2015 masters by four shots on sunday . the 21-year-old american led all week at the augusta national golf club . he shot final-round 70 to finish on 18 under par and take the green jacket . \\ \hline
   {\sc Drop + RE: } to all of my masters, \newline four sunday morning, american led alert \newline national golf club, final-round time \newline take a green jacket  \\ \hline
   {\sc Synonym + RE: } jordan, we are not the same, no masters! \newline four shots of the sun, the laughter \newline we were the most likely american led in a week \newline at the first club shot last finish, hey \newline get the green cap \\ \hline
   {\sc Shuffle + RE (\model{}): } masters, four shots on sunday \newline jordan, led me to the national club, the american way \newline golf week, green dine, par \newline finish my jacket, take my final-round start \\ \hlineB{2}
   {\sc Input (news): } the dallas native will play alongside justin rose in the final pairing . has set a scoring record for the first 54 holes of 16 under par . finished runner-up last year and is now determined to win . is first player since greg norman in 1996 to have lead after each round . \\ \hline
   {\sc Drop + RE: } dallas native play \newline i was born to be a slave \newline but now i'm on my own \newline and i'll be the first so \newline justin final scoring holes in par \newline last year determined to start \newline been a player, since greg \newline the only way to tell  \\ \hline
   {\sc Synonym + RE: } dallas, c4, i play with the same \newline g6, justin rose to the place \newline c1, ready to scoring the record \newline first holes in the firearm, then i remember \newline this is the first year \newline determined to win, first player, since marc ellen went here \\ \hline
   {\sc Shuffle + RE (\model{}): } justin rose, native gold \newline final par, scoring holes, set it off, play it again, justin rose \newline determined to win the first record, last year i was finished \newline greg player, he was a player from the beginning \newline since first i lead the worldball. \\ \hlineB{2}
    \end{tabular}
	\caption{Additional model outputs for style transfer from news articles to rap lyrics. }
	\label{tab:extra_examples_style_transfer_news}
\end{table*}

%% file: tables/abstract-lyrics.tex
\begin{table*}
    \centering
		\small 
		\begin{tabular}{p{14cm}} %{7.5cm}
    \hlineB{2} 
    [intro]

i am the oldest

the lyrics they just follow orders.

i am the oldest

the lyrics they just follow orders.

good trade-off of your style.

i am the oldest

the lyrics they just follow orders.

i rhyme more rhymes and moreover

move over I'm recording \newline

 [verse 1]

another verse written on the news of rap methods,

given to me in the form of an autoencoder

to develop the words that i rap, and i will be denoting

in my text, i am the only content,

i can be the same as an automatist,

i train rap lyrics to study different meaning when i approach words as i am,

I train lyrics that are the most definitive,

more essential than a scheme of three

more untouchable than an underflow

move over. pirana, the founder, moreover.

my rhyme lyrics are more than the rhyme over

(when i develop a verse) \newline

 [verse 2]

when i develop a verse i form a text from an art that is written on the news of an autoencoder rap

 another method given to a train that i have been through and i am not the only thing to do with

this is my reality

i will not be content with rap lyrics i approach with the meaning oh

my words are based on my attack.

my lyrics are essential as I generate rap.

my average rhyme scheme is to show you different content

in other words, i can't study my own admirations.

my raps are so amazing

the rhyme is paraphrasing. \newline

[bridge]

my results are very good like I'm a human being

my rap is in the convoy.

your lyrics will be so pre-dated.

(when i develop a verse) \newline

[outro]

I'm a human being

I'm a human being
    
   \\ \hline 
    \end{tabular}
	\caption{Lyrics of our demo song, described in Appendix \ref{sec:demo_song}.}
	\label{tab:abstract_lyrics}
\end{table*}